\documentclass[letterpaper, 10 pt, conference]{ieeeconf}
\usepackage{amsmath,amsfonts}
\usepackage{array}
\usepackage{textcomp}
\usepackage{stfloats}
\usepackage{url}
\usepackage{verbatim}
\usepackage{graphicx}
\usepackage{cite}
\usepackage{xcolor}
\usepackage{subcaption}
\usepackage{mathtools}
\usepackage{amssymb}
\usepackage{tabulary}
\usepackage{booktabs}
\usepackage[ruled,linesnumbered]{algorithm2e}
\usepackage{setspace}
\usepackage{multirow}
\usepackage{enumerate}
\makeatletter
\let\NAT@parse\undefined
\makeatother
\usepackage{hyperref}

\newcommand{\titlename}{ColAG}

\IEEEoverridecommandlockouts
\title{\LARGE \bf
\titlename: A Collaborative Air-Ground Framework\\for Perception-Limited UGVs' Navigation
}

\author{
  Zhehan Li\textsuperscript{$\dagger$ 1,2}, Rui Mao\textsuperscript{$\dagger$ 2,3}, Nanhe Chen\textsuperscript{1,2}, Chao Xu\textsuperscript{1,2}, Fei Gao\textsuperscript{1,2} and Yanjun Cao\textsuperscript{1,2}
  \thanks{\textsuperscript{$\dagger$} \textbf{Equal contribution.}}
  \thanks{This work was supported by National Nature Science Foundation of China under Grant No. 62103368. (Corresponding author: Yanjun Cao, Fei Gao.)}
  \thanks{\textsuperscript{1} State Key Laboratory of Industrial Control Technology, Institute of Cyber-Systems and Control, Zhejiang University, Hangzhou, 310027, China.}
  \thanks{\textsuperscript{2} Huzhou Institute of Zhejiang University, Huzhou, 313000, China.}
  \thanks{\textsuperscript{3} Dalian University of Technology, Dalian, 116024, China.}
  \thanks{E-mails:\tt\small \{zhehanli, nanhe\_chen, cxu, fgaoaa, yanjunhi\}@zju.edu.cn, speedforce@mail.dlut.edu.cn}
}

\makeatletter
\let\@oldmaketitle\@maketitle
\renewcommand{\@maketitle}{\@oldmaketitle
  \includegraphics[width=\textwidth]{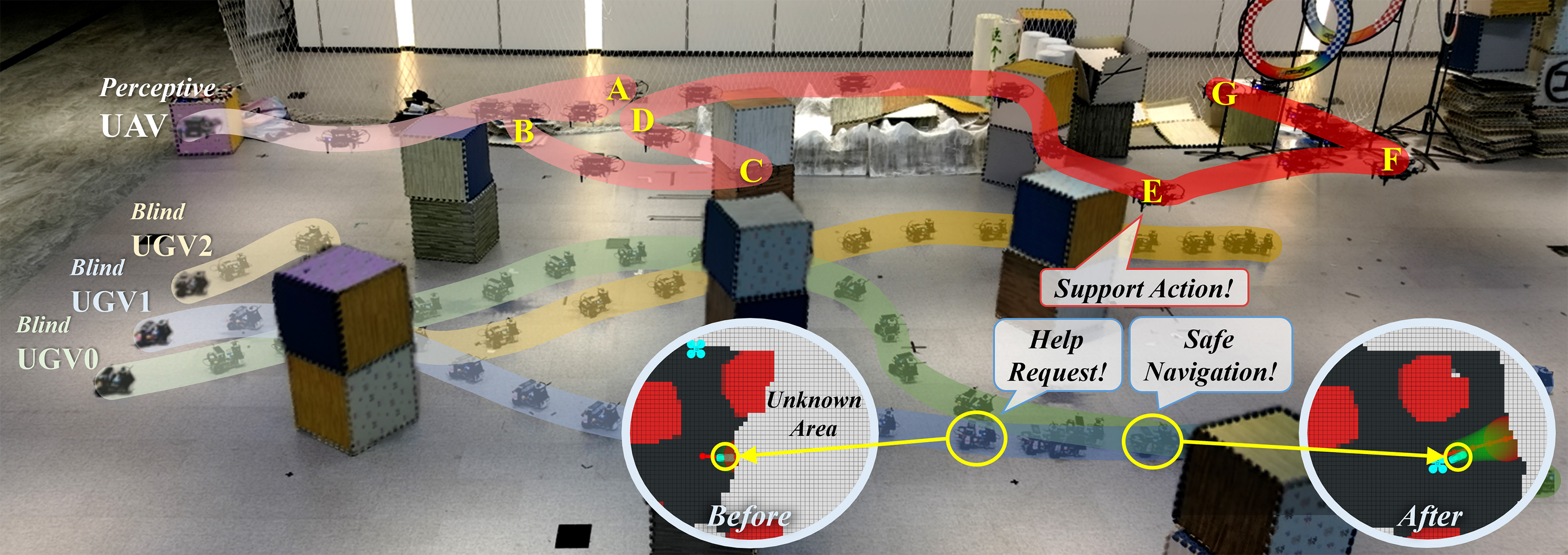}
  \captionsetup{font={small}}
  \captionof{figure}{Real-world experiment of \titlename~with a group of perception-limited UGVs and one fully functional UAV. The \titlename~system allows three blind UGVs to safely navigate through an area full of obstacles with help from the UAV. The UAV autonomously flies to points $A$-$G$ to support the UGVs avoiding potential collisions by improving state estimation accuracy and providing maps. $E$ for example, UGV1 requests help from the UAV because of its high state estimation uncertainty and potential to enter unknown terrains. After the arrival of the UAV, UGV1 can move forward safely.
}
  \label{fig: Top Figure}
}
\makeatother

\begin{document}

\maketitle

\begin{abstract}

  Perception is necessary for autonomous navigation in an unknown area crowded with obstacles.
  It's challenging for a robot to navigate safely without any sensors that can sense the environment, resulting in a \textit{blind} robot, and becomes more difficult when comes to a group of robots.
  However, it could be costly to equip all robots with expensive perception or SLAM systems.
  In this paper, we propose a novel system named \titlename, to solve the problem of autonomous navigation for a group of \textit{blind} UGVs by introducing cooperation with one UAV, which is the only robot that has full perception capabilities in the group.
  The UAV uses SLAM for its odometry and mapping while sharing this information with UGVs via limited relative pose estimation.
  The UGVs plan their trajectories in the received map and predict possible failures caused by the uncertainty of its wheel odometry and unknown risky areas.
  The UAV dynamically schedules waypoints to prevent UGVs from collisions, formulated as a Vehicle Routing Problem with Time Windows to optimize the UAV's trajectories and minimize time when UGVs have to wait to guarantee safety.
  We validate our system through extensive simulation with up to 7 UGVs and real-world experiments with 3 UGVs.

\end{abstract}

\section{Introduction}
Nowadays, unmanned ground vehicles (UGVs) and unmanned aerial vehicles (UAVs) are widely used in many fields, such as surveillance\cite{lin2018topology}, agriculture\cite{pretto2020building}, search-and-rescue\cite{delmerico2017active}\cite{miller2022stronger}, and transportation\cite{cognetti2014cooperative}\cite{guerin2015uav}, almost all relying on kinds of external sensors to sense the environment, such as LiDAR, camera, and radar etc.
Perception ability is one of the key functions for autonomous navigation in an unknown area crowded with obstacles, which is also the condition for the following navigation modules such as mapping, planning, and high-level strategy.
It's challenging for a perception-limited robot to navigate safely, especially if it lacks sensors to detect the environment, resulting in a \textit{blind} robot discussed in this paper.
Furthermore, it becomes more difficult when comes to a group of robots with potential inter-robot collision due to the lack of perception.
When the size of the group increases, it can be costly to equip all robots with expensive perception or SLAM systems.

In this paper, we target a problem of autonomous navigation for a group of low-cost {blind} UGVs, referred to as \textit{blind navigation}.
Our motivation derives from the fact that UGVs possess distinct advantages such as low cost, high load capacity, and low energy consumption.
UGVs are often used as mission execution units and work as a swarm due to these benefits, such as multi-robot collaborative transportation.
More than expensive to equip all robots with sensors, the on-ground viewpoint also can not make the best usage of these expensive sensors.
Whereas, UAVs excel in terms of exceptional mobility in 3D space and therefore have a much bigger perception area.
Combining their benefits to form a collaborative system comprising UGVs and UAVs can make an efficient system of demanding tasks at low cost.
Therefore, we introduce an aerial robot equipped with full perception capabilities, working as a shared remote mobile eye, cooperatively guiding a group of blind UGVs from the aerial view to solve this {blind navigation} problem.

Following this concept, the main problem of this system is how to use a perceptive UAV to guide perception-limited UGVs safely and efficiently in an unknown environment.
It is not trivial for one single UAV to guide multiple ground robots simultaneously.
The challenge mainly comes from three aspects, limitation of relative pose estimation (RPE), UGV's planning in a received map, and the scheduling of the UAV for dynamically changed situations.
In this system, the connections between UAV and UGVs are built from the relative state estimation system, such as AprilTag\cite{olson2011apriltag}, specially designed LED board\cite{yan2019active} or CREPES\cite{xun2023crepes}, etc.
Note that we do not consider the RPE techniques from global localization systems (e.g. motion capture systems) because of the dependency on infrastructure.
However, all these RPE techniques are direct observation and therefore limited by the field of view (FOV) and the occlusion of the environment.
One UAV can't guarantee that all UGVs are constantly viewed in its FOV for stable RPE.
The UGVs face the challenge of navigating in a shared map without direct environmental sensing.
Enabling the UGVs to achieve optimal planning, in a received map including free, obstacle and unknown areas, is crucial.
The treatment of unknown areas affects collision prediction as well as the planning quality dramatically.
The control of blind UGVs with only the wheel odometer faces substantial challenges if no further action is taken but just trajectory tracking.
The last point is about the planning of the UAV considering the dynamic nature of the UGVs group navigation.
The UAV needs to find the best planning trajectory considering the UGVs' changing situations to take action efficiently without wasting energy.

To overcome these difficulties and solve the {blind navigation} problem, we propose a novel Collaborative Air-Ground Framework (\titlename), which employs a perceptive UAV to guide multiple blind UGVs simultaneously.
The UAV is equipped with LiDAR and RPE device, while the UGVs are equipped with RPE device and wheel odometer only.
The UAV uses SLAM to obtain its odometry and map and shares them with UGVs to recover global consistent maps.
The UGVs receive the shared map and plan trajectory while using an EKF to estimate its odometry in the map.
The EKF fuses RPE measurements from the UAV and its wheel odometers.
The blind UGVs continuously predict possible collisions caused by the uncertainty of odometry estimation and unknown areas in the map.
Then the information of possible collisions is sent to the UAV and the UAV plans its motion to provide the UGVs with updated measurements to save them from collisions.
The contributions of this paper are summarized as follows:
\begin{enumerate}
  \item We propose \titlename, a novel framework to solve the problem of \textit{blind navigation} that a UAV guides a group of blind UGVs to achieve autonomous navigation in an unknown crowded environment.
  \item We design a path planning strategy for the blind UGVs, which aggressively plans the trajectory and conservatively predicts the possible collision of the UGVs caused by the localization uncertainty, to improve the efficiency and guarantee the safety of the UGVs.
  \item We formulate the scheduling of the UAV as dynamic vehicle routing problems with time windows (VRPTW) to provide the support order for UGVs' motion, which can minimize the total length of the UAV's trajectory and maximize the support before UGVs reach the predicted collision positions.
  \item We release our implementation of \titlename~for the reference and benefit of the community\footnote{\url{https://github.com/fast-fire/ColAG}}.
\end{enumerate}

\section{Related Works}

Considerable researches have studied heterogeneous robot systems, particularly in collaborative Air-Ground robotic systems.
A major amount of the researches focus on combining two kinds of robots to collaboratively carry out tasks such as exploration\cite{wang2020collaborative}\cite{butzke20153} and reconstruction\cite{qin2019autonomous}\cite{he2020ground}, leveraging the complementary strengths of each robot.
These researches aim to address the limitations inherent in different types of robots for overall performance and capabilities.

Some studies involve utilizing UAVs to assist UGVs in their movement, leveraging the UAV's broader FOV to provide enhanced environmental information to the UGVs.
Delmerico et al.\cite{delmerico2017active} and Peterson et al.\cite{peterson2018online} assemble imagery into an ortho mosaic and then classify it using classical computer vision methods to generate an initial trajectory for UGV.
Miller et al.\cite{miller2022stronger} use semantic information between aerial and ground realize RPE and run in real-time with multi-UGV.
This collaboration allows the UGVs to benefit from the UAV's aerial perspective and gather more comprehensive situational awareness.

The environment-aware sensors carried by the UGVs were removed in some studies, and the RPE between the UAV and the UGVs was used to obtain the localization of the UGVs.
Cognetti et al.\cite{cognetti2014cooperative} use one UAV with a camera to provide localization for multiple UGVs and develop a cooperative control scheme to keep the UGVs inside the camera field of view.
Guerin et al.\cite{guerin2015uav} use a visual servo on UAV to keep a leader UGV in the center of the image plane for UGV's localization, and the navigation waypoints are given by a human operator.
Mueggler et al.\cite{mueggler2014aerial} localize the UGV with AprilTag which can be detected by the UAV's camera, to compensate for the drift of the UGV's wheel odometry.
\cite{cognetti2014cooperative}\cite{guerin2015uav}\cite{mueggler2014aerial}\cite{shen2017collaborative} incorporate visual devices on UAVs to provide crucial localization information for the UGV, which need UAV's continuous observation.
This is not only inefficient but also limits the UAV's mobility, yet without the observation, the UGV's localization faces drifting and uncertainty.

Localization uncertainty brings collision risks to the system, which is studied in planning and control.
EKF is the most used uncertainty propagation method.
Zhu et al.\cite{zhu2019chance}, Kamel et al.\cite{kamel2017robust} and Patil et al.\cite{patil2012estimating} propagate uncertainties along the planned trajectory using EKF.
Different methods are used to check collision with uncertainty.
\cite{zhu2019chance} calculate approximated tight bound of collision chance, \cite{kamel2017robust} use the max eigenvalue of uncertainty matrix as $\sigma$, then use $3\sigma$ as collision distance, and \cite{patil2012estimating} truncate the a priori distribution to get collision probability.

Efficient scheduling of UAVs is necessary to guide these blind UGV systems, which has been studied in autonomous exploration problems.
Zhou et al.\cite{zhou2021fuel} and Meng et al.\cite{meng2017two} find a global tour for a UAV to visit viewpoints efficiently by formulating it as a Traveling Salesman Problem (TSP) with a little revision.
Zhou et al.\cite{zhou2023racer} and Gao et al.\cite{gao2022meeting} utilize a Constrained Vehicle Routing Problem (CVRP) to minimize the overall lengths of UAVs' path and balance the workloads for UAVs.
Stump et al.\cite{stump2011multi} consider a persistent surveillance problem, and cast it as a VRPTW to finding sequences of visits to discrete sites in a periodic fashion.

In \titlename, for UGV we design a strictly safe strategy for path planning, which includes trajectory generation and collision prediction based on uncertainty propagation, and for UAV we design a scheduling method formulated as VRPTW to guide UAV to support the UGVs.

\section{The \titlename~Framework}

\begin{figure}[h]
  \centering
  \includegraphics[width=0.48\textwidth]{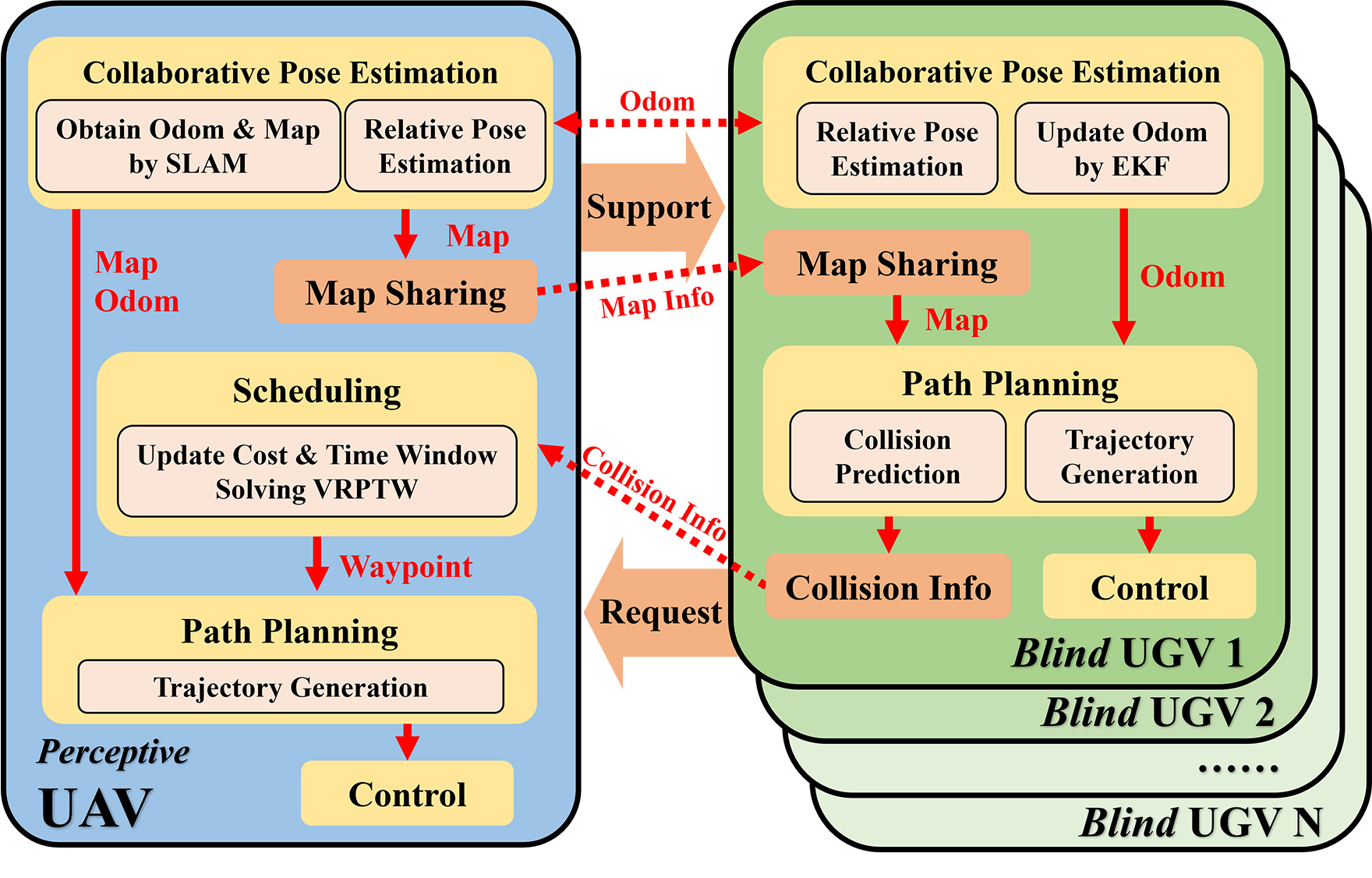}
  \captionsetup{font={small}}
  \caption{The system architecture of \titlename.}
  \label{fig: System Architecture}
\end{figure}

In \titlename~as shown in Fig. \ref{fig: System Architecture}, the relationship between UAV and blind UGVs like the relationship between helper and recipient.
The perceptive UAV runs SLAM for its map and odometry, supports UGVs collaborative pose estimation (\ref{sec: Collaborative Pose Estimation}) with RPE and shares the map (\ref{sec: Map Sharing}).
The blind UGVs plan trajectories in the received map and predict possible collisions (\ref{sec: Path Planning of UGV}), then send the collision information to the UAV to request support.
The UAV utilizes this information to dynamically update VRPTW to schedule its waypoints for the optimal support order of the UGVs (\ref{sec: Scheduling of UAV}).

\subsection{Collaborative Pose Estimation}
\label{sec: Collaborative Pose Estimation}

The UAV obtains its odometry and map by LiDAR SLAM directly, while the blind UGVs have to fuse its wheel odometry and RPE with UAV for accurate positioning.
To obtain global consistent pose estimation, the UGVs employ EKF to fuse RPE, UAV's odometry, and wheel odometers.
We write the state of the UGV at time $k$ as $\boldsymbol{x}_{k}$, including 2D position and heading.
\begin{equation}
  \begin{aligned}
    \boldsymbol{x}_{k} & = \begin{bmatrix}\boldsymbol{p}_{k} & \theta_{k}\end{bmatrix}^{T}
    = \begin{bmatrix}{p_{x}}_{k} & {p_{y}}_{k} & \theta_{k}\end{bmatrix}^{T} \\
  \end{aligned}
  \label{eq: ugv state}
\end{equation}

\subsubsection{Pridiction Model}

We use the wheel odometers of the UGV as the control input $\boldsymbol{u}_{k}$ of the UGV, including linear velocity and angular velocity.
\begin{equation}
  \begin{aligned}
    \boldsymbol{u}_{k} & = \begin{bmatrix}\boldsymbol{v}_{k} & \omega_{k}\end{bmatrix}^{T}
    = \begin{bmatrix}{v_{x}}_{k} & {v_{y}}_{k} & \omega_{k}\end{bmatrix}^{T} \\
  \end{aligned}
\end{equation}

The UGV state transition model can be written as
\begin{equation}
  \begin{aligned}
    \boldsymbol{x}_{k}       & = f(\boldsymbol{x}_{k-1}, \boldsymbol{u}_{k})
    = \begin{bmatrix}  \boldsymbol{p}_{k-1} + \boldsymbol{R}\{\theta_{k-1}\} \boldsymbol{v_{k}} \Delta t \\
        \theta_{k-1} + \omega_{k} \Delta t\end{bmatrix}^{T} \\
    \boldsymbol{R}\{\theta\} & = \begin{bmatrix}\cos\theta & -\sin\theta \\ \sin\theta & \cos\theta\end{bmatrix}                 \\
  \end{aligned}
  \label{eq: ugv model}
\end{equation}

Where $\boldsymbol{p}$ is the position of the UGV in the world frame, $\theta$ is the yaw angle of the UGV, $\boldsymbol{v}$ is the velocity of the UGV, $\omega$ is the angular velocity of the UGV, $\Delta t$ is the time interval of the prediction step, $\boldsymbol{R}\{\theta\}$ is the rotation matrix of the UGV.

The EKF prediction step is written as
\begin{equation}
  \begin{aligned}
    \hat{\boldsymbol{x}}_{k|k-1} & = f(\hat{\boldsymbol{x}}_{k-1|k-1}, \boldsymbol{u}_{k})                                                                         \\
    \boldsymbol{P}_{k|k-1}       & = \boldsymbol{F}_{k} \boldsymbol{P}_{k-1|k-1} \boldsymbol{F}_{k}^T + \boldsymbol{B}_{k} \boldsymbol{Q}_{k} \boldsymbol{B}_{k}^T \\
  \end{aligned}
  \label{eq: ekf prediction}
\end{equation}

\subsubsection{Measurement Model}

We use poses observed from the UAV as the measurement of the UGVs. This observed UGVs' poses can be obtained by multiplying the RPE and the odometry of the UAV. RPE can be obtained by the RPE device. The odometry of the UAV can be obtained by the SLAM algorithm running on the UAV.
Formulated as
\begin{equation}
  \begin{aligned}
    \boldsymbol{z}_{k} & = h(\boldsymbol{x}_{k})
    = \begin{bmatrix}
        \tilde{\boldsymbol{p}}_{k} &
        \tilde{\theta}_{k}
      \end{bmatrix}^{T}                                                             \\
                       & = \begin{bmatrix}
                             x\{{{}^W\boldsymbol{p}_{G}}_{k}\} &
                             y\{{{}^W\boldsymbol{p}_{G}}_{k}\} &
                             yaw\{{{}^W\boldsymbol{R}_{G}}_{k}\}
                           \end{bmatrix}^{T} \\
  \end{aligned}
\end{equation}

\begin{equation}
  \begin{aligned}
    {}^W\boldsymbol{R}_{G} & = {}^W\boldsymbol{R}_{A} {}^A\boldsymbol{R}_{G}                          \\
    {}^W\boldsymbol{p}_{G} & = {}^W\boldsymbol{p}_{A} + {}^W\boldsymbol{R}_{A} {}^A\boldsymbol{p}_{G} \\
  \end{aligned}
  \label{eq: observe}
\end{equation}

Where ${}^W\boldsymbol{R}_{G}$ and ${}^W\boldsymbol{p}_{G}$ are the rotation matrix and position of the UGV in world frame, ${}^W\boldsymbol{R}_{A}$ and ${}^W\boldsymbol{p}_{A}$ are the rotation matrix and position of the UAV in world frame (obtained by SLAM), ${}^A\boldsymbol{R}_{G}$ and ${}^A\boldsymbol{p}_{G}$ are the rotation matrix and position of the UGV in UAV frame (obtained by RPE).

When a certain UGV is first observed by the UAV, the EKF can be initialized by the RPE.
The EKF measurement step is written as
\begin{equation}
  \begin{aligned}
    \boldsymbol{K}_{k}         & = \boldsymbol{P}_{k|k-1} \boldsymbol{H}_{k}^T (\boldsymbol{H}_{k} \boldsymbol{P}_{k|k-1} \boldsymbol{H}_{k}^T + \boldsymbol{R}_{k})^{-1} \\
    \hat{\boldsymbol{x}}_{k|k} & = \boldsymbol{x}_{k|k-1} + \boldsymbol{K}_{k} (\boldsymbol{z}_{k} - h(\hat{\boldsymbol{x}}_{k|k-1}))                                     \\
    \boldsymbol{P}_{k|k}       & = (\boldsymbol{I} - \boldsymbol{K}_{k} \boldsymbol{H}_{k}) \boldsymbol{P}_{k|k-1}                                                        \\
  \end{aligned}
  \label{eq: ekf measurement}
\end{equation}

\subsection{Map Sharing}
\label{sec: Map Sharing}

We use the occupancy grid map as the map format.
In UAV, the grid map is a 3D array, each cell in the array represents a grid in the environment, and the value of the cell represents the state of the grid.
The state of the grid can be free, occupied, or unknown, depending on the value of the cell.
Note that these three states are treated differently on the UGVs for navigation, detailed in Sec. \ref{sec: Path Planning of UGV}.
To reduce the bandwidth needed for map sharing, the UAV only sends the addresses of occupied cells and unknown cells in a specific region around the UGVs, and the free cells can be inferred as the remaining cells in the region.
When the UGVs receive the map information, they store it in the same format as the UAV and merge the map according to the address of occupied cells and unknown cells.

\subsection{Path Planning of UGV}
\label{sec: Path Planning of UGV}

\begin{figure}[h]
  \centering
  \includegraphics[width=0.40\textwidth]{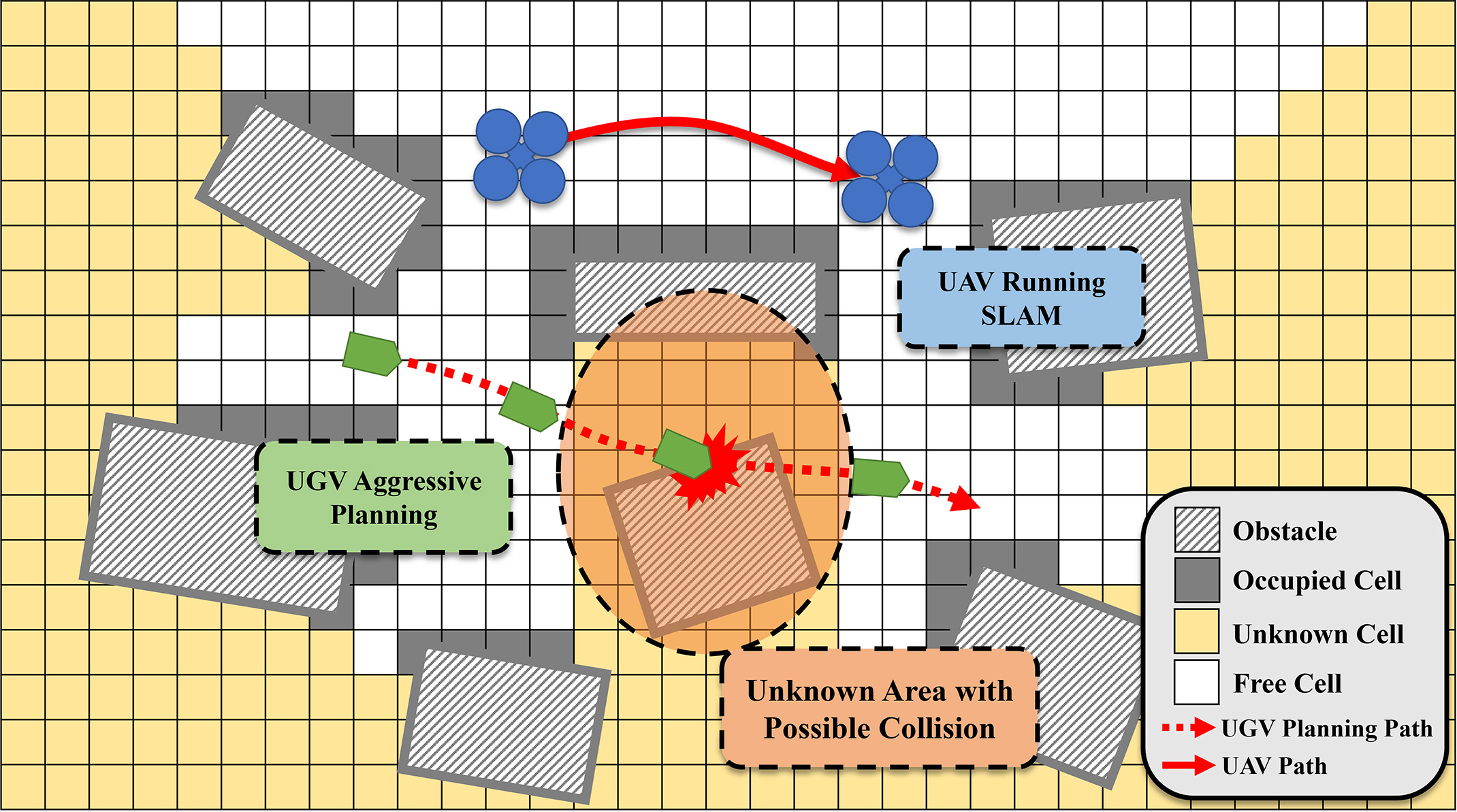}
  \captionsetup{font={small}}
  \caption{The map constructed by the UAV is not suitable for UGVs' planning because of the occlusion of the environment. If the UGVs assume the unknown cells are free, possible collisions exist along the planning trajectory.}
  \label{fig: Framework}
\end{figure}

Due to the different viewpoints of the UAV and the UGVs, the environment surrounding the UGVs may be occluded in the UAV's map, presented as unknown cells in UGVs' map.
These unknown cells in the reality can be free or occupied for UGVs, causing the risk of collision between UGVs and obstacles, shown in Fig. \ref{fig: Framework}.

\subsubsection{Trajectory Generation in P-Map}
\begin{figure}[h]
  \centering
  \includegraphics[width=0.48\textwidth]{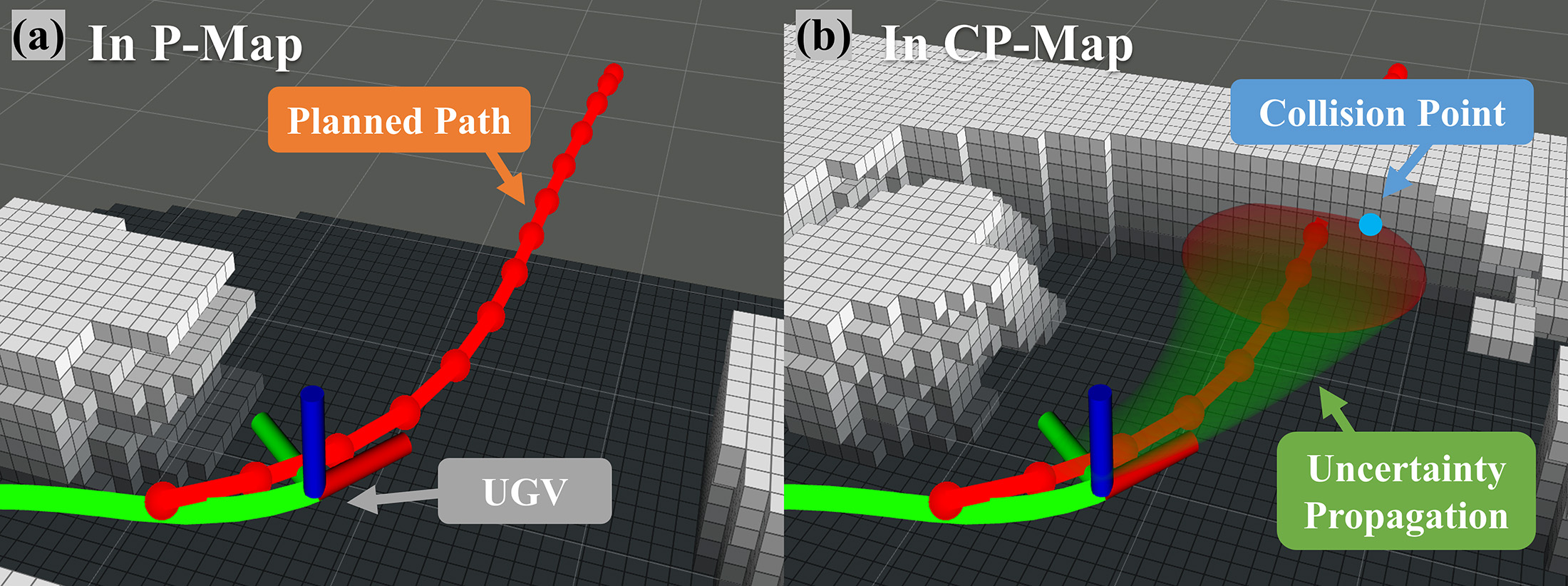}
  \captionsetup{font={small}}
  \caption{The path planning of UGVs. (a) The UGV plans in P-Map, the red line is the planned path. (b) The UGV propagates uncertainty and predicts collision in CP-Map along the trajectory, the growing green area is the predicted covariance ellipses, the red ellipse is the collided ellipse, and the blue point is the collided point.}
  \label{fig: Collision Prediction}
\end{figure}

It is conservative to suggest that the UGVs can only move in the known-free environment when planning, which may reduce the efficiency of the system.
We adopt an aggressive strategy that the UGVs plan in the planning map (P-Map) which assumes the unknown cells are free, as shown in Fig. \ref{fig: Collision Prediction}(a).
This approach aligns with conventional trajectory generation techniques utilized in autonomous navigation\cite{zhou2021ego}.

\subsubsection{Potential Risk of Collision}
\label{sec: Potential Risk of Collision}
This aggressive strategy brings collision risk to the UGVs, we consider two situations:
First, when RPE observation from UAV is delayed than expected, the UGVs' odometers will shift, increasing the pose uncertainty.
Second, obstacles might exist in unknown cells that the UGVs cannot perceive.

\subsubsection{Uncertainty Propagation of UGVs}
We use the covariance matrix of the EKF to represent the uncertainty of the pose estimation of UGVs.
We calculate the prediction steps by Eq. \ref{eq: ekf prediction} along the planning trajectory in a certain distance to propagate the UGV's uncertainty.

\subsubsection{Uncertainty-Based Collision Detect in CP-Map}
To deal with the two situations in \ref{sec: Potential Risk of Collision} uniformly, we propose an uncertainty-based collision prediction method in the collision predicting map (CP-Map) which suggests the unknown cells as occupied, as shown in Fig. \ref{fig: Collision Prediction}(b).

Since the covariance matrix of the EKF is an approximate Gaussian distribution, we can use it to find the possible collision of the UGVs in the CP-Map.
Due to the $3\sigma$-rule in two-dimensional Gaussian distribution, that the probability of a random variable being within $3\sigma$ range of its mean is 98.89\%, we can assume that the UGVs will be safe if it keeps within $3\sigma$ of its pose estimation along with the trajectory.
The calculation is as follows:
\begin{itemize}
  \item First, we calculate the eigen values and eigen vectors of the covariance matrix of the EKF
        \begin{equation}
          \begin{aligned}
             & \boldsymbol{Q}^{-1} \boldsymbol{P}_{xy} \boldsymbol{Q} = \boldsymbol{\Lambda} = diag(\lambda_1, \lambda_2) \\
             & \boldsymbol{Q} = \begin{bmatrix}\boldsymbol{v}_1\;\boldsymbol{v}_2\;\end{bmatrix}                          \\
          \end{aligned}
          \label{eq: eigen}
        \end{equation}

        Where $\boldsymbol{P}_{xy}$ is the covariance matrix of the EKF in XY plane, $\boldsymbol{\Lambda}$ is the eigen values of $\boldsymbol{P}_{xy}$, $\boldsymbol{v}_1$ and $\boldsymbol{v}_2$ are the eigen vectors of $\boldsymbol{P}_{xy}$.

  \item Then we calculate the $3\sigma$ ellipse
        \begin{equation}
          \begin{aligned}
            a & = 3\sqrt{\lambda_1} , \; b = 3\sqrt{\lambda_2}                         \\
            1 & \geq \frac{((\boldsymbol{p}-\boldsymbol{p}_G)\boldsymbol{v}_1)^2}{a^2}
            + \frac{((\boldsymbol{p}-\boldsymbol{p}_G)\boldsymbol{v}_2)^2}{b^2}        \\
          \end{aligned}
          \label{eq: ellipse}
        \end{equation}

        Where $a$ and $b$ are the length of the semi-major axis and semi-minor axis of the ellipse, $\boldsymbol{p}$ is the position of the cell in the ellipse, $\boldsymbol{p}_G$ is the position of the UGV.

  \item Finally, we traverse the cells occupied by the ellipses along the planned trajectory and find the center of collision ellipse $\boldsymbol{p}_{pc}$ and collision time $t_{pc}$ of the UGVs.
        As LiDAR has some blind spots, we choose an expected support point $\boldsymbol{p}_{ps}$ with a certain distance back from the $\boldsymbol{p}_{pc}$ in the direction of planned velocity.
        $\boldsymbol{p}_{ps}$ and $t_{pc}$ are combined as the \textbf{collision information}.
\end{itemize}

Note that UGV will stop and wait for UAV when collision time is less than a threshold $t_c$, which is $0.4\;s$ in our work.

\subsection{Scheduling of UAV}
\label{sec: Scheduling of UAV}

\begin{figure}[h]
  \centering
  \includegraphics[width=0.40\textwidth]{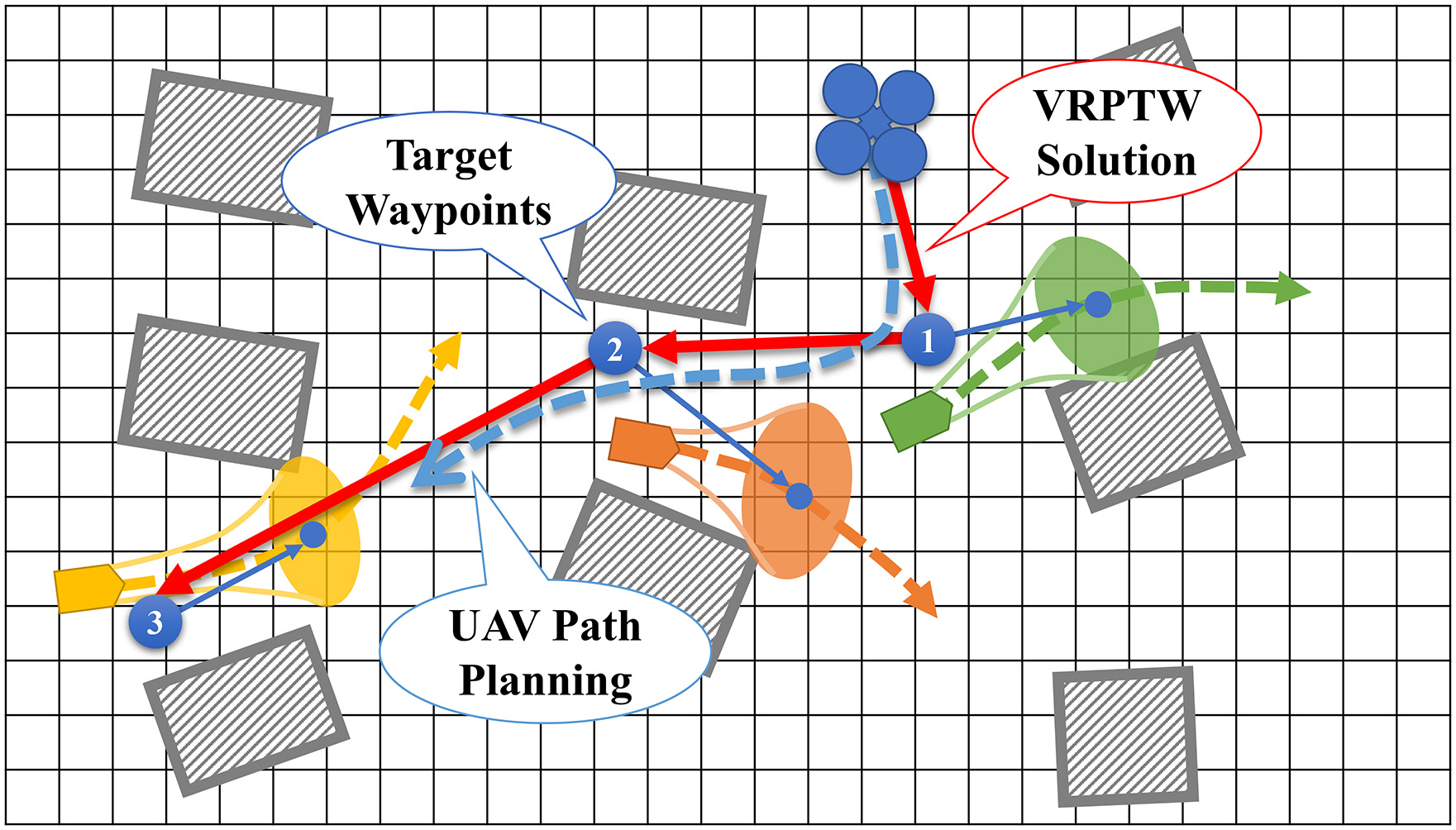}
  \captionsetup{font={small}}
  \caption{The scheduling of UAV, the VRPTW takes the predicted collision information and gets the solution of the order to support the UGVs, which are used for UAV's trajectory planning.}
  \label{fig: Scheduling}
\end{figure}

Since we only have the UAV to sense the environment, we need to efficiently dispatch the UAV to help the UGVs maximize their efficiency.
When the UAV comes to support, it can provide RPE measurement and nearby map information for the UGVs, so that the UGVs can update their pose estimation, obtain the information of potential collision cells, and replan the trajectory to avoid collision.
To minimize the UGVs' waiting time because of possible collision, we need the UAV to support the UGVs before the UGVs reach the potential collision cells, shown in Fig. \ref{fig: Scheduling}.

We formulate the scheduling problem of UAV as a VRPTW by designing the time cost matrix $\boldsymbol{M}_{vrptw}$ and the time window matrix $\boldsymbol{V}_{vrptw}$.
Assume there are $N$ collision points, $\boldsymbol{M}_{vrptw}$ corresponds to a $N+1$ dimensions square matrix and $\boldsymbol{T}_{vrptw}$ is a $(N+1)\times2$ matrix, computed as:
\begin{equation}
  \begin{aligned}
     & \boldsymbol{M}_{vrptw}(k_1,k_2)=\boldsymbol{M}_{vrptw}(k_2,k_1)                                     \\
     & =t_b({\boldsymbol{p}_{ps}}_{k_1},{\boldsymbol{p}_{ps}}_{k_2}),\;k_1,k_2\in\{1,2,\dots,N\}           \\
     & \boldsymbol{M}_{vrptw}(0,k)=t_b(\boldsymbol{p}_{A},{\boldsymbol{p}_{ps}}_{k}),\;k\in\{1,2,\dots,N\} \\
     & \boldsymbol{M}_{vrptw}(k,0)=0                                                                       \\
  \end{aligned}
  \label{eq: time matrix}
\end{equation}
\begin{equation}
  \begin{aligned}
     & \boldsymbol{T}_{vrptw}(k,0)=0                                  \\
     & \boldsymbol{T}_{vrptw}(k,1)={t_{pc}}_{k},\;k\in\{1,2,\dots,N\} \\
  \end{aligned}
  \label{eq: time window}
\end{equation}

Where $\boldsymbol{p}_{A}$ is the position of the UAV. $t_b({\boldsymbol{p}_{ps}}_{k_1},{\boldsymbol{p}_{ps}}_{k_2})$ is the time cost from ${\boldsymbol{p}_{ps}}_{k_1}$ to ${\boldsymbol{p}_{ps}}_{k_2}$, calculated by
\begin{equation}
  \begin{aligned}
    t_b({\boldsymbol{p}_{ps}}_{k_1},{\boldsymbol{p}_{ps}}_{k_2})=\frac{\|{\boldsymbol{p}_{ps}}_{k_1}-{\boldsymbol{p}_{ps}}_{k_2}\|}{v_{max}}
  \end{aligned}
  \label{eq: time cost}
\end{equation}

We consider the continuity of the UAV's speed and calculate the time cost by
\begin{equation}
  \begin{aligned}
     & \boldsymbol{p}_{dis}                               =
    \boldsymbol{p}_{A}-{\boldsymbol{p}_{ps}}_{k},
    v_{tan}                                             =
    \boldsymbol{p}_{dis}/\|\boldsymbol{p}_{dis}\|\cdot\boldsymbol{v}_{A} \\
     & dis_a                                              =
    ({v_{max}}^2 - {v_{d}}^2)/(2{a_{max}})                               \\
     & t_b(\boldsymbol{p}_{A},{\boldsymbol{p}_{ps}}_{k})  =              \\
     & \left\{
    \begin{aligned}
       & \frac{\sqrt{v_{tan}^2+2a_{max}dis_a}-v_{tan}}{a_{max}},
       & dis_a >  \|\boldsymbol{p}_{dis}\|                                               \\
       & \frac{v_{max}-v_{tan}}{a_{max}}+\frac{\|\boldsymbol{p}_{dis}\|-dis_a}{v_{max}},
       & dis_a\leq\|\boldsymbol{p}_{dis}\|                                               \\
    \end{aligned}
    \right.
  \end{aligned}
  \label{eq: time cost begin}
\end{equation}

We use OR-Tools\cite{ortools} to solve the scheduling problem.
The formulation of VRPTW is to ensure that the UAV can support the UGVs before the UGVs reach the possible collision position.
However, there may be no solution to the VRPTW problem, which means that the UAV cannot support all UGVs in time.
In this case, we gradually increase the $v_{max}$ until the VRPTW problem can find a solution.

\section{Experiment}

To verify the effectiveness of \titlename, we conduct simulation with different configurations and real-world experiments.
EGO-Swarm\cite{zhou2021ego} is used to generate smooth trajectories of the UAV and UGVs.

\subsection{Simulation Experiment}

In the simulation experiment, we use the MARSIM\cite{kong2023marsim} simulator to simulate the quadrotor UAV and LiDAR, and use numerical simulation to simulate the UGVs with differential chassis.
We simulate the RPE by calculating true RPE and add it with Gaussian noise.

\begin{figure}[h]
  \centering
  \includegraphics[width=0.48\textwidth]{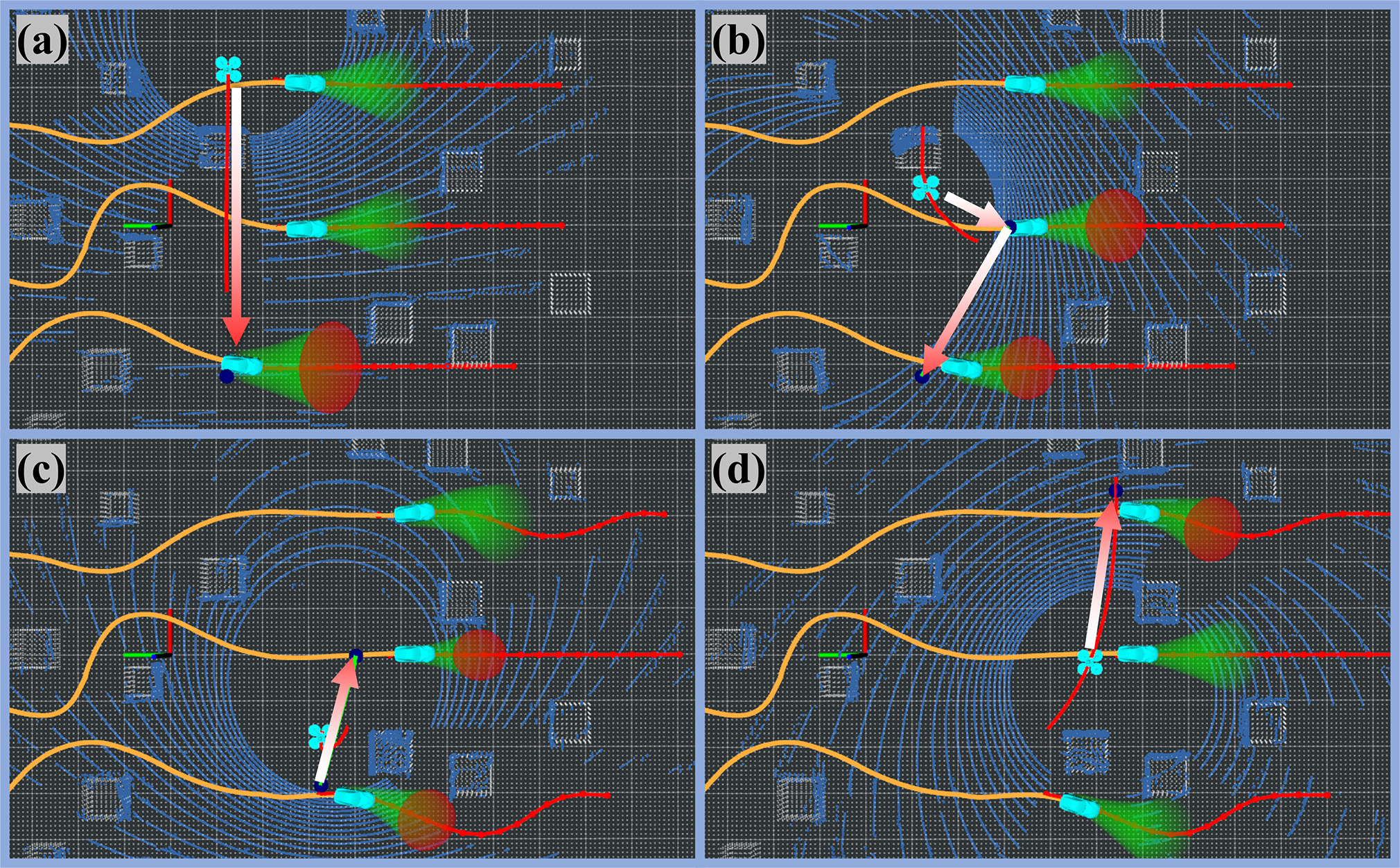}
  \captionsetup{font={small}}
  \caption{The \titlename\;simulation. In (a)-(d), the UAV supports the UGVs which are predicted to collide in the order of the VRPTW solution, providing RPE and LiDAR perception of the environment.}
  \label{fig: Simulation}
\end{figure}

The UGVs' dynimics limits are set to $\boldsymbol{v}_{max}=0.5\;m/s$ and $\boldsymbol{a}_{max}=1.0\;m/s^2$, and the UAV's are set to $\boldsymbol{v}_{max}=3.0\;m/s$ and $\boldsymbol{a}_{max}=1.0\;m/s^2$.
The simulated Gaussian noise of UGV's wheel odometers is set to $\boldsymbol{\sigma}(v_x)=0.0336\;m/s$, and $\boldsymbol{\sigma}(\omega)=0.0292\;rad/s$, which is measured on our real platform, and noise of RPE is set to $0.2\;m$ and $0.05\;rad$.
The sensor range of the LiDAR and RPE is set to $8m$ and $5m$, and the map-sharing range is set to $2.5m$.

We test the performance in sparse and dense environments with different numbers of UGVs.
The environments are $27\times27\times3\;m^3$ space with randomly generated obstacles, 40 in sparse environment and 80 in dense environment.
We evaluate the performance by the following metrics:
\begin{itemize}
  \item \textbf{UGV Reach Time (s):} The average time when UGV reaches the goal.
  \item \textbf{UGV Waiting Time (s):} The average time of UGV waiting because the predicted collision time is too small.
  \item \textbf{UGV Trajectory Length (m):} The average length of UGV's trajectory.
  \item \textbf{UAV Trajectory Length (m):} The average length of UAV's trajectory.
\end{itemize}

To prove our effectiveness, we also simulate UGVs with full perception, with no need for the UAV in the system.
\begin{itemize}
  \item \textbf{Self-Perception UGV Reach Time (s):} The average time when UGV reaches the goal, where UGVs are equipped with LiDAR and run SLAM independently.
  \item \textbf{Self-Perception UGV Trajectory Length (m):} The average length of UGV's trajectory, where UGVs are equipped with LiDAR and run SLAM independently.
\end{itemize}

\begin{table}[h]
  \centering
  \captionsetup{font={small}}
  \caption{\titlename\;Benchmark}
  \label{tab: Benchmark}
  \resizebox{\columnwidth}{!}{
    \begin{tabular}{|c|c|c|c|c|c|c|c|}
      \hline
      Envi.                                                        &
      \begin{tabular}[c]{@{}c@{}}UGV\\Num.\end{tabular}            &
      \begin{tabular}[c]{@{}c@{}}UGV\\Reach Time\end{tabular}      &
      \begin{tabular}[c]{@{}c@{}}UGV\\Wait. Time\end{tabular}      &
      \begin{tabular}[c]{@{}c@{}}UGV\\Traj. Len.\end{tabular}      &
      \begin{tabular}[c]{@{}c@{}}UAV\\Traj. Len.\end{tabular}      &
      \begin{tabular}[c]{@{}c@{}}S.P. UGV\\Reach Time\end{tabular} &
      \begin{tabular}[c]{@{}c@{}}S.P. UGV\\Traj. Len.\end{tabular}   \\
      \hline
      \multirow{4}{*}{Spar.}                                       &
      1                                                            &
      106.78                                                       &
      0.00                                                         &
      33.02                                                        &
      36.35                                                        &
      106.20                                                       &
      32.92                                                          \\
      \cline{2-8}
                                                                   &
      3                                                            &
      106.13                                                       &
      0.11                                                         &
      32.96                                                        &
      120.94                                                       &
      106.59                                                       &
      32.98                                                          \\
      \cline{2-8}
                                                                   &
      5                                                            &
      116.62                                                       &
      3.87                                                         &
      33.65                                                        &
      214.07                                                       &
      107.59                                                       &
      33.32                                                          \\
      \cline{2-8}
                                                                   &
      7                                                            &
      129.12                                                       &
      12.59                                                        &
      33.59                                                        &
      265.50                                                       &
      106.92                                                       &
      33.22                                                          \\
      \hline
      \multirow{4}{*}{Dense}                                       &
      1                                                            &
      112.30                                                       &
      0.00                                                         &
      34.95                                                        &
      38.27                                                        &
      111.04                                                       &
      34.52                                                          \\
      \cline{2-8}
                                                                   &
      3                                                            &
      112.38                                                       &
      0.07                                                         &
      34.73                                                        &
      133.08                                                       &
      110.69                                                       &
      34.53                                                          \\
      \cline{2-8}
                                                                   &
      5                                                            &
      119.54                                                       &
      4.86                                                         &
      34.68                                                        &
      216.05                                                       &
      110.50                                                       &
      34.38                                                          \\
      \cline{2-8}
                                                                   &
      7                                                            &
      143.18                                                       &
      24.65                                                        &
      34.98                                                        &
      304.34                                                       &
      111.88                                                       &
      34.94                                                          \\
      \hline
    \end{tabular}
  }
\end{table}
The simulation results are shown in Table \ref{tab: Benchmark}.
According to the results, \titlename~successfully guides the UGVs to cross sparse and dense environments, with different numbers of UGVs.
When the number of UGVs is less than 5, the reach time is close to the self-perception UGVs' reach time, and the UGVs' waiting time is small.
When the number of UGVs is greater than 5, the waiting time of the UGVs and the length of the UAV's trajectory increase, which is acceptable as the limitation of UAV.

To the best of our knowledge, there is no other work solved the {blind navigation} problem.
To verify the necessity of the parts in \titlename, we designed an ablation study with the following methods

\begin{itemize}
  \item \textbf{without RPE Measurement:} The UGVs only use RPE to initialize the EKF and run on their wheel odometers.
  \item \textbf{without Uncertainty Propagation:} The UGVs predict collision without considering the uncertainty.
  \item \textbf{without Scheduling:} The UAV will support the UGV with the smallest predicted collision time first.
  \item \textbf{without Time Window:} The UAV doesn't consider the time window of the possible collision, and the VRPTW formation degenerates to VRP.
\end{itemize}

\begin{table}[h]
  \centering
  \captionsetup{font={small}}
  \caption{\titlename\;Ablation}
  \label{tab: Comparsion}
  \resizebox{\columnwidth}{!}{
    \begin{tabular}{|c|c|c|c|c|c|}
      \hline
      Methods                                                      &
      \begin{tabular}[c]{@{}c@{}}UGV\\Reach Time\end{tabular}      &
      \begin{tabular}[c]{@{}c@{}}UGV Avg.\\Wait. Time\end{tabular} &
      \begin{tabular}[c]{@{}c@{}}UGV Max.\\Wait. Time\end{tabular} &
      \begin{tabular}[c]{@{}c@{}}UGV\\Traj. Len.\end{tabular}      &
      \begin{tabular}[c]{@{}c@{}}UAV\\Traj. Len.\end{tabular}        \\
      \hline
      Proposed                                                     &
      119.54                                                       &
      4.86                                                         &
      11.08                                                        &
      34.68                                                        &
      216.05                                                         \\
      \hline
      w/o RPE Meas.                                                &
      -                                                            &
      -                                                            &
      -                                                            &
      -                                                            &
      -                                                              \\
      \hline
      w/o Unce. Prop.                                              &
      -                                                            &
      -                                                            &
      -                                                            &
      -                                                            &
      -                                                              \\
      \hline
      w/o Sche.                                                    &
      135.73                                                       &
      18.10                                                        &
      27.85                                                        &
      34.85                                                        &
      239.71                                                         \\
      \hline
      w/o Time Wind.                                               &
      132.43                                                       &
      18.29                                                        &
      46.65                                                        &
      34.81                                                        &
      238.56                                                         \\
      \hline
    \end{tabular}
  }
\end{table}

We conducted the ablation study in dense environment with 5 UGVs and results are shown in Table \ref{tab: Comparsion}
Without RPE measurement and uncertainty propagation, the UGVs pose estimation drift and collide with the obstacles, resulting in mission failure.
Without scheduling considering flight time cost, the UAV can't plan an optimal global trajectory, leading to a longer trajectory length, and then can't support UGVs in time.
Without time window, the UAV always follows one UGV, leading to a much longer waiting time of other UGVs.

\subsection{Real-World Experiment}
\begin{figure}[h]
  \centering
  \includegraphics[width=0.48\textwidth]{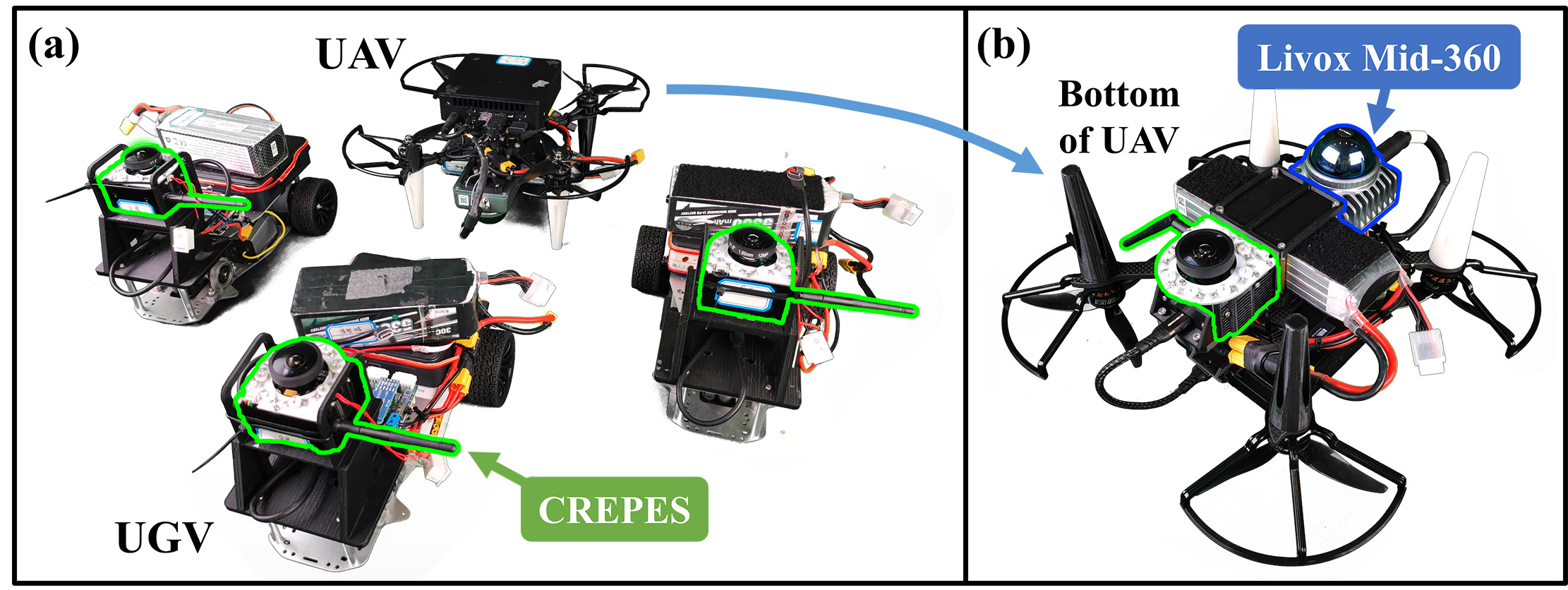}
  \captionsetup{font={small}}
  \caption{The UAV and UGVs platform, the green edge device is CREPES and the blue edge device is Livox Mid-360. (a) The UAV and 3 UGVs. (b) The bottom of the UAV.}
  \label{fig: Real World Platform}
\end{figure}

To validate the proposed method, we conduct real-world experiments. Fig. \ref{fig: Real World Platform} shows an quadrotor and three simple differential-driven UGVs we used.
All robots are equipped with CREPES\cite{xun2023crepes} as RPE and Intel NUC (i7-1165G7 CPU) as the computation platform.
The UAV is equipped with a Livox Mid-360 and runs Fast-lio\cite{xu2021fast} as its SLAM module.

The UGVs' dynimics limits are set to $\boldsymbol{v}_{max}=1.0\;m/s$ and $\boldsymbol{a}_{max}=1.0\;m/s^2$, and the UAV's are set to $\boldsymbol{v}_{max}=1.5\;m/s$ and $\boldsymbol{a}_{max}=0.5\;m/s^2$.
The range of the RPE is set to $3m$.
The other parameters are the same as the simulation.

\begin{figure}[h]
  \centering
  \includegraphics[width=0.48\textwidth]{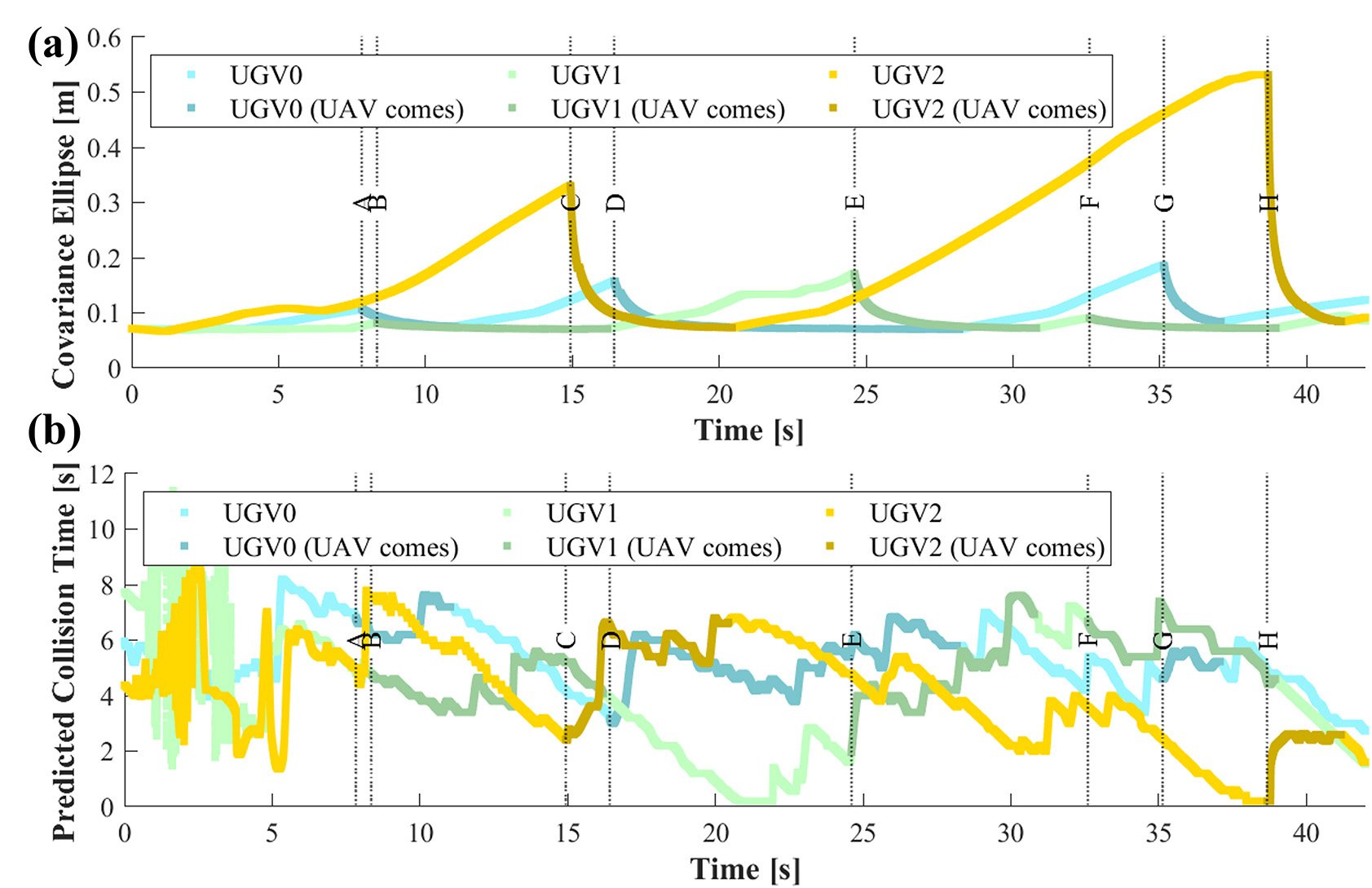}
  \captionsetup{font={small}}
  \caption{Real world experiment. $A$-$H$ are the points UAV came to help in Fig. \ref{fig: Top Figure}. (a) The propagation of the UGVs' state estimation uncertainty (half-long axis $\sqrt{\lambda_1}$ of ellipse). (b) The UGVs' predicted collision time.}
  \label{fig: Real World Experiment}
\end{figure}

The real-world experiment is shown in Fig. \ref{fig: Top Figure}.
The UAV and UGVs are placed in a $20\times10\times3\;m^3$ arena and the goal is set to $11m$ away from the start point.
In Fig. \ref{fig: Real World Experiment}, as long as the UAV comes to support, the uncertainty of the UGVs' pose estimation is decreased, and the predicted collision time is increased, guaranteeing the safety of the UGVs.
The result in Table \ref{tab: Real World Experiment} shows that \titlename\; performed well with the direct RPE device CREPES, the UGVs reached the goal without collision and the UAV supported the UGVs in time.

\begin{table}[h]
  \centering
  \captionsetup{font={small}}
  \caption{Real World Experiment}
  \label{tab: Real World Experiment}
  \resizebox{0.7\columnwidth}{!}{
    \begin{tabular}{|c|c|c|c|c|}
      \hline
      \begin{tabular}[c]{@{}c@{}}UGV\\ID\end{tabular}         &
      \begin{tabular}[c]{@{}c@{}}UGV\\Reach Time\end{tabular} &
      \begin{tabular}[c]{@{}c@{}}UGV\\Wait. Time\end{tabular} &
      \begin{tabular}[c]{@{}c@{}}UGV\\Traj. Len.\end{tabular} &
      \begin{tabular}[c]{@{}c@{}}UAV\\Traj. Len.\end{tabular}   \\
      \hline
      0                                                       &
      41.15                                                   &
      0.00                                                    &
      12.92                                                   &
      \multirow{4}{*}{25.75}                                    \\
      \cline{1-4}
      1                                                       &
      38.60                                                   &
      1.20                                                    &
      11.07                                                   &
      \\
      \cline{1-4}
      2                                                       &
      42.75                                                   &
      0.40                                                    &
      13.52                                                   &
      \\
      \cline{1-4}
      Avg.                                                    &
      40.83                                                   &
      0.53                                                    &
      12.50                                                   &
      \\
      \hline
    \end{tabular}
  }
\end{table}

\section{Conclusion}

In this work, we propose a novel cooperative air-ground framework to solve the \textit{blind navigation} problem in which only one perceptive UAV guides a group of blind UGVs to navigate in unknown environments.
A path planning strategy is introduced to consider the uncertainty of UGVs' pose estimation and the unknown parts of the map, guaranteeing the UGVs safety.
Based on collision prediction, a customized VRPTW scheduling strategy is proposed for the UAV to support the UGVs, optimizing the UAV's trajectory and the UGVs' waiting time.
Simulation and real-world experiments are conducted to verify the effectiveness of the proposed method.
In the future, we will consider the initiative exploration of the UAV on the possible collision trajectory, and the cooperation of multiple UAVs to support the UGVs.

\bibliographystyle{IEEEtran}
\bibliography{root}

\end{document}